\documentclass[a4paper, USenglish, cleveref, autoref, thm-restate]{lipics-v2021}

\title{SAT-based Circuit Local Improvement} 

\author{Alexander~S. Kulikov}{Steklov Mathematical Institute at St.~Petersburg, Russian Academy of Sciences \and St.~Petersburg State University \and \url{https://logic.pdmi.ras.ru/~kulikov/}}{kulikov@logic.pdmi.ras.ru}{https://orcid.org/0000-0002-5656-0336}{}

\author{Danila Pechenev}{St.~Petersburg State University}{danil-pechenev@mail.ru}{https://orcid.org/0000-0003-0575-0807}{}

\author{Nikita Slezkin}{St.~Petersburg State University}{ne.slezkin@gmail.com}{https://orcid.org/0000-0003-1904-9261}{}

\authorrunning{A.\,S.~Kulikov, D.~Pechenev, N.~Slezkin} 

\Copyright{Alexander~S. Kulikov, Danila Pechenev, and Nikita Slezkin} 



\relatedversion{} 
\relatedversiondetails[linktext={ECCC TR21-044}, cite={DBLP:journals/eccc/KulikovS21}]{Technical report}{https://eccc.weizmann.ac.il/report/2021/044/} 

\supplement{Source code:}
\supplementdetails[linktext={GitHub}]{repository}{https://github.com/alexanderskulikov/circuit_improvement} 

\nolinenumbers 

\hideLIPIcs  

\EventEditors{John Q. Open and Joan R. Access}
\EventNoEds{2}
\EventLongTitle{42nd Conference on Very Important Topics (CVIT 2016)}
\EventShortTitle{CVIT 2016}
\EventAcronym{CVIT}
\EventYear{2016}
\EventDate{December 24--27, 2016}
\EventLocation{Little Whinging, United Kingdom}
\EventLogo{}
\SeriesVolume{42}
\ArticleNo{23}

\usepackage{tikz}
\usetikzlibrary{math}

\tikzstyle{input}=[draw=none, inner sep=.2mm]
\tikzstyle{gate}=[draw, circle, inner sep=.2mm, minimum size=4mm]
\tikzstyle{l}=[draw=none, rectangle, fill=white, inner sep=.4mm]

\newenvironment{mypic}{
  \begin{center}
  \begin{tikzpicture}
}{
  \end{tikzpicture}
  \end{center}
}

\usepackage{booktabs}
\usepackage[color=red!5,colorinlistoftodos]{todonotes}
\usepackage{url}
\usepackage{booktabs}

\usepackage[frozencache]{minted} 
\setminted{bgcolor=gray!10}

\DeclareMathOperator{\SUM}{SUM}
\DeclareMathOperator{\MOD}{MOD}
\DeclareMathOperator{\THR}{THR}
\DeclareMathOperator{\size}{size}

\DeclareMathOperator{\IN}{IN}
\DeclareMathOperator{\MID}{MID}
\DeclareMathOperator{\OUT}{OUT}
\DeclareMathOperator{\MDFA}{MDFA}

\begin{document}
    \sloppy
\maketitle

\begin{abstract}
Finding exact circuit size
is a~notorious optimization
problem in practice. Whereas modern computers
and algorithmic techniques allow to~find a~circuit
of~size seven in~blink of an~eye, it~may take more
than a~week to~search for a~circuit of~size thirteen.
One of the reasons of~this behavior is~that the search
space is~enormous: the number of~circuits of size~$s$
is~$s^{\Theta(s)}$, the number of~Boolean functions on~$n$ variables is~$2^{2^n}$.

In~this paper, we~explore the following natural
heuristic idea for decreasing the size of
a~given circuit: go~through all its subcircuits
of moderate size and check whether
any of~them can be~improved by~reducing to~SAT.
This may be viewed
as a~local search approach: we~search for a~smaller
circuit in a~ball around a~given circuit.
Through this approach, we~prove
new upper bounds on the circuit size
of~various symmetric functions. We~also demonstrate that
some upper bounds that were~proved by~hand
decades ago, nowadays can be~found automatically
in a~few seconds.
\end{abstract}


\section{Boolean Circuits}
\subsection{Exact Circuit Synthesis}

A~Boolean \emph{straight line program}
of size~$r$ for input variables $(x_1, \dotsc, x_n)$
is a~sequence of~$r$~instructions where each
instruction $g \gets h \circ k$
applies a~binary Boolean operation~$\circ$ to
two operands $h,k$ each of which is either an~input bit
or the result of a~previous instruction.
If $m$~instructions are designated as outputs,
the straight line program computes a~function
$\{0,1\}^n \to \{0,1\}^m$ in a~natural way.
We denote the set of all such functions by $B_{n,m}$ and we let $B_n=B_{n,1}$.
For
a~Boolean function $f \colon \{0,1\}^n \to \{0,1\}^m$,
by $\size(f)$ we denote the minimum size of
a~straight line program
computing~$f$. A~Boolean \emph{circuit}
shows a~flow graph of a~program.

\begin{figure}[!ht]
\begin{minipage}{.28\textwidth}
\inputminted[firstline=15,lastline=18]{python}{../tutorial.py}
\end{minipage}
\begin{minipage}{.18\textwidth}
\begin{tikzpicture}[label distance=-.9mm,scale=.9]
\foreach \n/\x/\y in {1/0/1, 2/1/1}
  \node[input] (x\n) at (\x, \y) {$x_{\n}$};
\node[gate, label=left:$w_1$] (g1) at (0,0) {$\land$};
\node[gate, label=right:$w_0$] (g2) at (1,0) {$\oplus$};
\foreach \f/\t in {x1/g1, x1/g2, x2/g1, x2/g2}
  \draw[->] (\f) -- (\t);
\end{tikzpicture}
\end{minipage}
\begin{minipage}{.33\textwidth}
\inputminted[firstline=22,lastline=28]{python}{../tutorial.py}
\end{minipage}
\begin{minipage}{.18\textwidth}
~
\begin{tikzpicture}[label distance=-.9mm,scale=.9]
\foreach \n/\x/\y in {1/0/3, 2/1/3, 3/2/3}
  \node[input] (x\n) at (\x, \y) {$x_{\n}$};
\node[gate,label=left:$a$] (g1) at (0.5,2) {$\oplus$};
\node[gate,label=left:$b$] (g2) at (1.5,2) {$\oplus$};
\node[gate,label=left:$c$] (g3) at (0.5,1) {$\lor$};
\node[gate, label=right:$w_0$] (g4) at (1.5,1) {$\oplus$};
\node[gate, label=right:$w_1$] (g5) at (0.5,0) {$\oplus$};
\foreach \f/\t in {x1/g1, x2/g1, x2/g2, x3/g2, g1/g3, g2/g3, g1/g4, g3/g5, g4/g5}
  \draw[->] (\f) -- (\t);
\path (x3) edge[bend left,->] (g4);
\end{tikzpicture}
\end{minipage}
\caption{Optimal size straight line programs and circuits for $\SUM_2$ and $\SUM_3$. These two circuits are known as~\emph{half adder} and \emph{full adder}.}
\label{figure:sum23}
\end{figure}
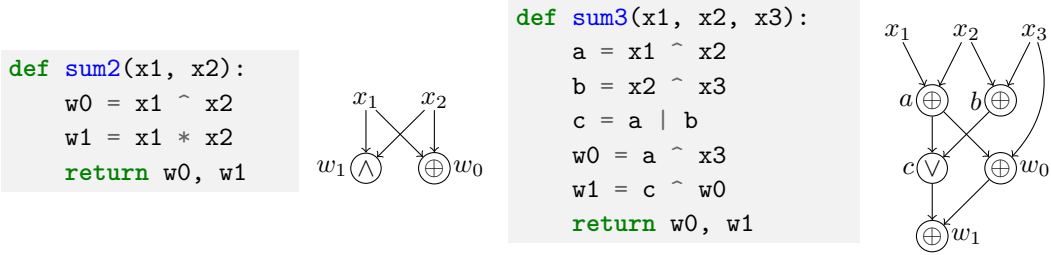

Figure~\ref{figure:sum23} gives an~example for
the
$\SUM_n \colon \{0,1\}^n \to \{0,1\}^l$ function
that computes the binary representation of~the sum of~$n$~bits:
\[\SUM_n(x_1, \dotsc, x_n)=(w_0, w_1, \dotsc, w_{l-1})\colon \sum_{i=1}^{n}x_i=\sum_{i=0}^{l-1}2^iw_i \text{, \, where } l=\lceil \log_2(n+1)\rceil \, .\]
This function transforms $n$~bits
of weight~0 into $l$~bits
of~weights $(0,1,\dotsc,l-1)$.
The straight line
programs are given
in~\texttt{Python} programming language
so~that
it~is particularly easy to~verify their correctness.
For example, the program for $\SUM_3$ can be verified
with just three lines of code:
\inputminted[firstline=32,lastline=36]{python}{../tutorial.py}

Determining $\size(f)$ requires
proving lower bounds:
to~show that $\size(f)> \alpha$,
one needs to~prove that \emph{every} circuit
of~size at~most~$\alpha$ does not compute~$f$.
Known lower bounds are far from being satisfactory:
the strongest known lower bound for a~function family
in~NP is~$(3+1/86)n-o(n)$~\cite{DBLP:conf/focs/FindGHK16}.
Here, by a~\emph{function family} we mean an~infinite sequence of~functions $\{f_n\}_{n=1}^{\infty}$ where $f_n \in B_n$.

Even proving lower bounds for specific functions (rather than function families) is~difficult. Brute force approaches become impractical quickly: $|B_n|=2^{2^n}$, hence already for $n=6$, one cannot just enumerate all
functions from~$B_n$; also, the number of~circuits
of~size~$s$ is $s^{\Theta(s)}$, hence checking all circuits of size~$s$ takes reasonable time for small values of~$s$ only.
Knuth~\cite{Knuth:2008:ACP:1377542} found the exact circuit size of all functions from~$B_4$ and~$B_5$.

Finding the exact value of $\size(f)$ for $f \in B_6$
is~already a~difficult computational task for modern
computers and techniques. One approach
is~to~translate a~statement ``there exists a~circuit
of~size~$s$ computing~$f$'' to a~Boolean formula and
to~pass~it to~a~SAT-solver. Then, if the formula
is~satisfiable, one decodes a~circuit from its satisfying
assignment; otherwise, one gets a~(computer generated) proof of a~lower bound $\size(f) > s$. This circuit synthesis approach was proposed
by~Kojevnikov et al.~\cite{DBLP:conf/sat/KojevnikovKY09}
and, since then, has been used in various circuit synthesis programs (\texttt{abc}~\cite{abc}, \texttt{mockturtle}~\cite{EPFLLibraries}, \texttt{sat-chains}~\cite{knuthreduction}).

The state-of-the-art SAT-solvers are surprisingly efficient and allow to~handle
various practically important problems (with millions
of~variables) and even help to~resolve open
problems~\cite{DBLP:conf/cade/BrakensiekHMN20}.
Still, already for small values of $n$~and~$s$ the problem
of~finding a~circuit of~size~$s$ for a~function from~$B_n$
is~difficult for SAT-solvers.
We demonstrate the limits
of~this approach on~\emph{counting} functions:
\begin{equation}\label{eq:mod}
    \MOD_n^{m,r}(x_1, \dotsc, x_n)=[x_1+\dotsb+x_n \equiv r \bmod m]
\end{equation}
(here, $[\cdot]$ is the Iverson bracket: $[S]$~is equal to~$1$ if $S$~is true and is equal to~$0$ otherwise).
Using SAT-solvers, Knuth~\cite[solution to exercise~$480$]{Knuth:2015:ACP:2898950}
found $\size(\MOD_n^{3,r})$ for all $3 \le n \le 5$ and all $0 \le r \le 2$:
\begin{center}
    \begin{tabular}{lccc}
        \toprule
              & $r=0$ & $r=1$ & $r=2$\\
        \midrule
        $n=3$ & $3$ & $4$ & $4$\\
        $n=4$ & $7$ & $7$ & $6$\\
        $n=5$ & $10$ & $9$ & $10$\\
        \bottomrule
    \end{tabular}
\end{center}
Generalizing this pattern, he~made the following conjecture:
\begin{equation}\label{conjecture}
\size(\MOD_n^{3,r})=3n-5-[(n+r) \equiv 0\bmod 3] \text{ for all $n \ge 3$ and $r$.}
\end{equation}
He was also able to~prove (using SAT-solvers)
that $\size(\MOD_6^{3,0})=12$
and wrote:
``The case $n=6$ and $r \neq 0$, which lies tantalizingly close to the limits of
today's solvers, is still unknown.''

To~summarize, our current abilities for checking whether there exists a~Boolean circuit of~size~$s$ are roughly the following:
\begin{itemize}
\item for $s \le 6$, this can be~done in a~few seconds;
\item for $7 \le s \le 12$, this can (sometimes)
be~done in a~few days;
\item for $s \ge 13$, this is~out of~reach.
\end{itemize}

\subsection{New Results}
In~this paper, we~explore the limits of~the following natural idea: given a~circuit, try to~improve its size by~improving (using SAT-solvers, for example) the size
of~its subcircuit of~size seven. This is a~kind of a~local search approach: we~have no~possibility to~go through the whole space of all circuits, but we can at~least
search in a~neighborhood of a~given circuit.
This allows~us to~work with circuits consisting
of many gates.

As the results of experiments, we~show several circuits
for which the approach described above leads to improved upper bounds.
\begin{itemize}
    \item We support Knuth's conjecture~\eqref{conjecture} for $\MOD_n^{3,r}$ by~proving the matching upper bound:
    \[\size(\MOD_n^{3,r}) \le 3n-5-[(n+r) \equiv 0\bmod 3] \text{ for all $n \ge 3$ and $r$.}\]
    This improves slightly the previously known upper bound $\size(\MOD_n^{3,r}) \le 3n-4$ by~Demenkov et al.~\cite{DBLP:journals/ipl/DemenkovKKY10}.
    \item We~present improvements for $\size(\SUM_n)$ for various small~$n$ and show that some of~these circuits and their parts can be~used as~building blocks to~design efficient circuits for other functions in~semiautomatic fashion. In~particular, we~show that a~part of an~optimal circuit for $\SUM_5$ can be~used to~build optimal circuits of~size~$2.5n$ for $\MOD_n^{4,r}$~\cite{DBLP:journals/mst/Stockmeyer77} and best known circuits of~size~$4.5n+o(n)$ for $\SUM_n$~\cite{DBLP:journals/ipl/DemenkovKKY10}.
    In~turn, an~efficient circuit for $\SUM_5$ can be~found
    in a~few seconds if~one starts from a~standard circuit for $\SUM_5$ composed
    out of~two full adders and one half adder.
    \item We~design new circuits for the threshold function defined as~follows:
    \[\THR_n^k(x_1, \dotsc, x_n)=[x_1+\dotsb+x_n \ge k] \, .\]
    The best known upper bounds for $\THR$ are the following:
    \begin{align*}
        \size(\THR_n^k) &\le kn+o(n) \text{ for $2 \le k \le 4$~\cite{Dunne84} (see also~\cite[6.2, Theorem~2.3]{DBLP:books/teu/Wegener87}),}\\
        \size(\THR_n^k) &\le 4.5n+o(n) \text{ for $5 \le k$~\cite{DBLP:journals/ipl/DemenkovKKY10}.}
    \end{align*}
    We~get the following improvement:
    \begin{align*}
        \size(\THR_n^k) &\le (4.5-2^{2-\lceil\log_2k\rceil})n+o(n) \text{ for $4 \le k = O(1)$.}
    \end{align*}
    In~particular, $\size(\THR_n^4) \le 3.5n+o(n)$ and $\size(\THR_n^k) \le 4n+o(n)$ for $5 \le k \le 8$.
\end{itemize}
The improved upper bounds are obtained in a~semiautomatic fashion:
\begin{enumerate}
    \item first, we~automatically improve a~given small circuit with a~fixed number of~inputs using SAT-solvers;
    \item then, we~generalize~it to~every input size.
\end{enumerate}
For some function families,
the second step is~already known (for example, given a~small circuit
for $\SUM_5$, it is not difficult to~use~it as a~building block
to~design an~efficient circuit for $\SUM_n$ for every~$n$; see Section~\ref{section:sum}), though in~general this still needs to~be
done manually.

\subsection{Related work}
The approach we~use in this paper follows the SAT-based
local improvement method (SLIM): to~improve an~existing
discrete structure one goes through all its substructures
of~size accessible to a~SAT-solver.
SLIM has been applied successfully to~the following
structures:
branchwidth~\cite{DBLP:journals/tocl/LodhaOS19},
treewidth~\cite{DBLP:conf/sat/FichteLS17},
treedepth~\cite{DBLP:conf/cp/RamaswamyS20},
Bayesian network structure learning~\cite{DBLP:conf/aaai/RamaswamyS21},
decision tree learning~\cite{DBLP:conf/aaai/SchidlerS21}.

\section{Program: Feature Overview and Evaluation}
The program is~implemented in~\texttt{Python}.
We give a~high-level overview of~its main features below.
All the code shown below
can be~found in~the
file \texttt{tutorial.py} at~\cite{git-improvement}.
One may run~it after installing a~few \texttt{Python} modules. Alternatively, one may run the Jupyter notebook
\texttt{tutorial.ipynb} in~the cloud (without installing
anything) by pressing the badge ``Colab''
at~the repository page~\cite{git-improvement}.

\subsection{Manipulating Circuits}
\begin{figure}[t]
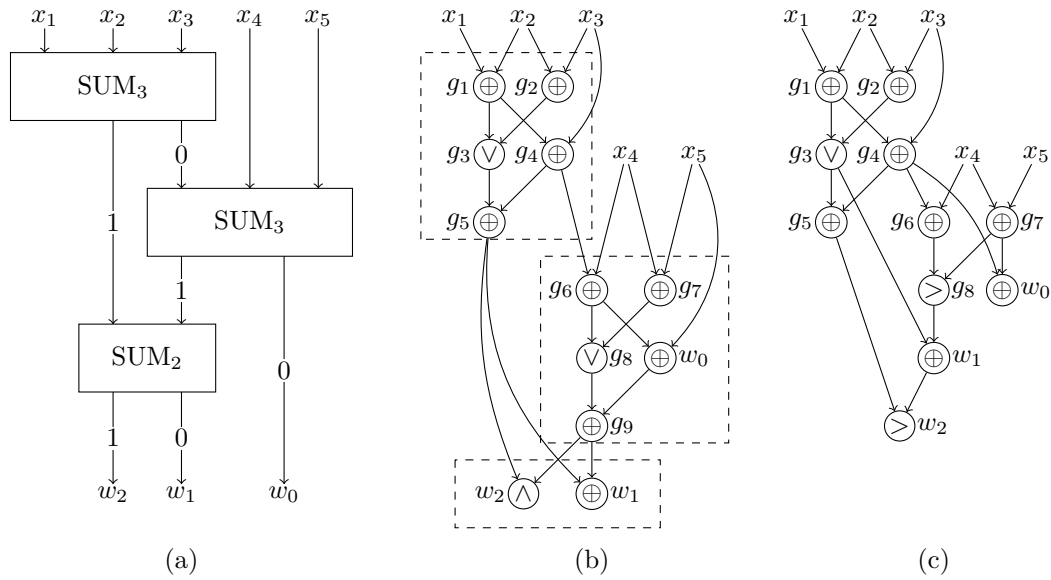

\begin{mypic}
\begin{scope}[scale=.9]
\begin{scope}[yshift=-10mm]
\foreach \n in {1,...,5}
  \node[input] (\n) at (\n,6) {$x_{\n}$};
\draw (0.5, 5.5) rectangle (3.5, 4.5); \node at (2, 5) {$\SUM_3$};
\foreach \n in {1, 2, 3}
  \draw[->] (\n) -- (\n, 5.5);
\draw (2.5, 3.5) rectangle (5.5, 2.5); \node at (4, 3) {$\SUM_3$};
\path (3, 4.5) edge[->] node[l] {0} (3, 3.5);
\foreach \n in {4, 5}
  \draw[->] (\n) -- (\n, 3.5);
\draw (1.5, 1.5) rectangle (3.5, 0.5); \node at (2.5, 1) {$\SUM_2$};
\path (2, 4.5) edge[->] node[l] {1} (2, 1.5);
\path (3, 2.5) edge[->] node[l] {1} (3, 1.5);
\node[input] (w2) at (2,-1) {$w_2$};
\node[input] (w1) at (3,-1) {$w_1$};
\node[input] (w0) at (4.5,-1) {$w_0$};
\path (2, 0.5) edge[->] node[l] {1} (w2);
\path (3, 0.5) edge[->] node[l] {0} (w1);
\path (4.5, 2.5) edge[->] node[l] {0} (w0);
\end{scope}

\begin{scope}[label distance=-1mm, xshift=70mm, yshift=20mm]
\foreach \n/\x/\y in {1/0/3, 2/1/3, 3/2/3, 4/2.5/1, 5/3.5/1}
  \node[input] (x\n) at (\x, \y) {$x_{\n}$};
\node[gate,label=left:$g_1$] (g1) at (0.5,2) {$\oplus$};
\node[gate,label=left:$g_2$] (g2) at (1.5,2) {$\oplus$};
\node[gate,label=left:$g_3$] (g3) at (0.5,1) {$\lor$};
\node[gate,label=left:$g_4$] (g4) at (1.5,1) {$\oplus$};
\node[gate,label=left:$g_5$] (g5) at (0.5,0) {$\oplus$};
\node[gate,label=left:$g_6$] (g6) at (2,-1) {$\oplus$};
\node[gate,label=right:$g_7$] (g7) at (3,-1) {$\oplus$};
\node[gate,label=right:$g_8$] (g8) at (2,-2) {$\lor$};
\node[gate, label=right:$w_0$] (g9) at (3,-2) {$\oplus$};
\node[gate, label=right:$g_9$] (g10) at (2,-3) {$\oplus$};
\node[gate, label=right:$w_1$] (g11) at (2,-4) {$\oplus$};
\node[gate, label=left:$w_2$] (g12) at (1,-4) {$\land$};

\foreach \f/\t in {x1/g1, x2/g1, x2/g2, x3/g2, g1/g3, g2/g3, g1/g4, g3/g5, g4/g5, g4/g6, x4/g6, x4/g7, x5/g7, g6/g8, g7/g8, g8/g10, g6/g9, g9/g10, g10/g11, g10/g12}
  \draw[->] (\f) -- (\t);

\path (x3) edge[->,bend left] (g4);
\path (x5) edge[->,bend left=35] (g9);
\path (g5) edge[->,bend right=25] (g11);
\path (g5) edge[->,bend right=15] (g12);

\draw[dashed] (-0.5,-0.25) rectangle (2,2.5);
\draw[dashed] (1.25,-3.25) rectangle (4,-0.5);
\draw[dashed] (0,-3.5) rectangle (3,-4.5);
\end{scope}

\begin{scope}[label distance=-1mm, xshift=120mm, yshift=20mm]
\foreach \n/\x/\y in {1/0/3, 2/1/3, 3/2/3, 4/2.5/1, 5/3.5/1}
  \node[input] (x\n) at (\x, \y) {$x_{\n}$};
\node[gate,label=left:$g_1$] (g1) at (0.5,2) {$\oplus$};
\node[gate,label=left:$g_2$] (g2) at (1.5,2) {$\oplus$};
\node[gate,label=left:$g_3$] (g3) at (0.5,1) {$\lor$};
\node[gate,label=left:$g_4$] (g4) at (1.5,1) {$\oplus$};
\node[gate,label=left:$g_5$] (g5) at (0.5,0) {$\oplus$};
\node[gate,label=left:$g_6$] (g6) at (2,0) {$\oplus$};
\node[gate,label=right:$g_7$] (g7) at (3,0) {$\oplus$};
\node[gate,label=right:$g_8$] (g8) at (2,-1) {$>$};
\node[gate, label=right:$w_0$] (g9) at (3,-1) {$\oplus$};
\node[gate, label=right:$w_1$] (g10) at (2,-2) {$\oplus$};
\node[gate, label=right:$w_2$] (g11) at (1.5,-3) {$>$};

\foreach \f/\t in {x1/g1, x2/g1, x2/g2, x3/g2, g1/g3, g2/g3, g1/g4, g3/g5, g4/g5, x4/g6, g4/g6, x4/g7, x5/g7, g6/g8, g7/g8, g7/g9, g3/g10, g8/g10, g10/g11, g5/g11}
  \draw[->] (\f) -- (\t);

\path (x3) edge[->,bend left] (g4);
\path (g4) edge[->,bend left=20] (g9);
\end{scope}

\foreach \x/\n in {3/a, 9/b, 14/c}
  \node at (\x,-3) {(\n)};
\end{scope}
\end{mypic}
\caption{(a)~A~schematic circuit for $\SUM_5$ composed out of two full adders and one half adder. (b)~The corresponding circuit of size~$12$. (c)~An~improved circuit of size~$11$.}
\label{figure:sumfive}
\end{figure}

This is done through the \mintinline{python}{Circuit}
class. One can load and save circuits as~well~as
print and draw them. A~nicely looking layout of
a~circuit is produced by the \texttt{pygraphviz} module~\cite{pygraphviz}. The program also contains some built-in
circuits that can be used as~building blocks.
The following sample code constructs a~circuit
for $\SUM_5$ out of two full adders and
one half adder. This construction is~shown
in~Figure~\ref{figure:sumfive}(a). Then,
the circuit is verified via the
\mintinline{python}{check_sum_circuit} method.
Finally, the circuit is drawn. As a~result, one gets
a~picture similar to~the one in~Figure~\ref{figure:sumfive}(b).

\inputminted[firstline=40,lastline=47]{python}{../tutorial.py}

\subsection{Finding Efficient Circuits}
The class \mintinline{python}{CircuitFinder}
allows to~check whether there exists a~circuit
of~the required
size for a~given Boolean function. For example,
one may discover the full adder as follows. (The function
\mintinline{python}{sum_n} returns the list of $\lceil \log_2(n+1) \rceil$ bits of~the binary representation of~the sum of~$n$~bits.)

\inputminted[firstline=51,lastline=58]{python}{../tutorial.py}

This is~done by~encoding the task as~a~CNF formula
and invoking a~SAT-solver
(via the \texttt{pysat} module~\cite{DBLP:conf/sat/IgnatievMM18}).
The reduction to~SAT is~described
in~\cite{DBLP:conf/sat/KojevnikovKY09}. Basically, one translates
a~statement ``there exists a~circuit of~size~$s$ comuting a~given function~$f \colon \{0,1\}^n \to \{0,1\}^m$'' to CNF-SAT. To~do this, one introduces many
auxiliary variables: for example, for every $x \in \{0,1\}^n$ and every $1 \le i \le r$, one uses a~variable that is~responsible for the value of the $i$-th gate
on an~input~$x$.

As~mentioned in~the introduction, the limits
of~applicability of~this approach (for finding a~circuit
of~size~$s$) are roughly the following:
for $s \le 6$, it~usually works in~less than a~minute;
for $7 \le s \le 12$, it~may already take up~to
several hours or days; for $s \ge 13$, it~becomes almost impractical. The running time may vary a~lot
for inputs of the same length. In~particular,
it usually takes much longer to~prove that
the required circuit does not exist (by~proving that the corresponding formula is~unsatisfiable). Table~\ref{table:runningtimes} reports the running time
of~this approach on several datasets.

\begin{table}[!ht]
\begin{center}
\begin{tabular}{lllr}
\toprule
function & circuit size & status & time (sec.)\\
\midrule
$\SUM_5$ & 12 & SAT & 141.4\\
$\SUM_5$ & 11 & SAT & 337.8\\
$\MOD_4^{3,0}$ & $7$ & SAT & 0.2\\
$\MOD_4^{3,0}$ & $6$ & UNSAT & 1178.8\\
$\MOD_4^{3,1}$ & $7$ & SAT & 0.2\\
$\MOD_4^{3,1}$ & $6$ & UNSAT & 1756.5\\
$\MOD_4^{3,2}$ & $6$ & SAT & 0.2\\
$\MOD_4^{3,2}$ & $5$ & UNSAT & 12.6\\
$\MOD_5^{3,0}$ & $10$ & SAT & 90.1\\
$\MOD_5^{3,1}$ & $9$ & SAT & 50.1\\
$\MOD_5^{3,2}$ & $10$ & SAT & 74.3\\
\bottomrule
\end{tabular}
\end{center}
\caption{The running time of \texttt{CircuitFinder} on~various Boolean functions.} \label{table:runningtimes}
\end{table}

\subsection{Improving Circuits}
The method \mintinline{python}{improve_circuit}
goes through all subcircuits of a~given size
of a~given circuit and checks whether any
of~them can be~replaced by a~smaller subcircuit
(computing the same function) via \mintinline{python}{find_circuit}. For example, applying this method
to~the circuit from Figure~\ref{figure:sumfive}(b)
gives the circuit from Figure~\ref{figure:sumfive}(c)
in~a~few seconds.
\inputminted[firstline=62,lastline=67]{python}{../tutorial.py}
Table~\ref{table:improvementrunningtimes} shows the time
taken by~\texttt{improve\_circuit} to~improve some
of~the known circuits for $\SUM$, $\MOD^3$, and $\THR^4$.
For $\SUM$, we~start from known circuits of size about $5n$
(composed out of~full adders and half adders). For $\MOD^3$,
we start from circuits of size $3n-4$ presented by~Demenkov et al.~\cite{DBLP:journals/ipl/DemenkovKKY10}. For $\THR^4$, we start from circuits of
size about~$5n$ (the start by~computing $\SUM_n$ and then compare the
resulting $\log n$-bit integer to~$4$).

\begin{table}[ht]
\begin{center}
\begin{tabular}{llr}
\toprule
function & circuit size & time (sec.)\\
\midrule
$\SUM_5$ & $12 \to 11$ & 6.7\\
$\SUM_7$ & $20 \to 19$ & 5.8\\
\midrule
$\MOD_6^{3,0}$ & $15 \to 14$ & 17.0\\
$\MOD_6^{3,1}$ & $15 \to 14$ & 17.2\\
$\MOD_6^{3,2}$ & $14 \to 13$ & 16.7\\
$\MOD_7^{3,0}$ & $17 \to 16$ & 31.3\\
$\MOD_7^{3,1}$ & $17 \to 16$ & 33.6\\
$\MOD_7^{3,2}$ & $16 \to 15$ & 30.5\\
\midrule
$\THR^4_5$ & $23 \to 10$ & $38.6$\\
$\THR^4_6$ & $28 \to 14$ & $42.1$\\
$\THR^4_7$ & $31 \to 17$ & $43.8$\\
$\THR^4_8$ & $40 \to 22$ & $55.1$\\
\bottomrule
\end{tabular}
\end{center}
\caption{The running time of \texttt{improve\_circuit} on~various Boolean functions.} \label{table:improvementrunningtimes}
\end{table}

\section{New Circuits}
In~this section, we~present new circuits for symmetric functions found with the help of~the program. A~function $f(x_1,\dotsc,x_n)$ is called \emph{symmetric} if~its value depends on~$\sum_{i=1}^nx_i$ only. They are among the most basic Boolean functions:
\begin{itemize}
\item to~specify an~arbitrary Boolean function
from~$B_n$, one needs
to~write down its truth table of length~$2^n$; symmetric functions allow for more compact representation: it is enough to specify $n+1$ bits (for each of $n+1$ values
of~$\sum_{i=1}^nx_i$);
\item circuit complexity of almost all functions
of~$n$ variables is exponential ($\Theta(2^n/n)$), whereas any symmetric function can~be computed by a~linear size circuit ($O(n)$).
\end{itemize}
Despite simplicity of~symmetric functions, we still do~not
know how optimal circuits look like for most of~them. Below, we~present new circuits for some of these functions.

\subsection{Sum Function}\label{section:sum}
The $\SUM$ function is a~fundamental symmetric function: for any symmetric $f \in B_n$, $\size(f)\le \size(\SUM_n)+o(n)$. The reason for this is that any function from~$B_n$ can be~computed by a~circuit of size $O(2^n/n)$ by the results of Muller~\cite{M56} and Lupanov~\cite{Lup59}. This allows to compute any symmetric $f(x_1, \dotsc, x_n) \in B_n$ as follows: first, compute $\SUM_n(x_1, \dotsc, x_n)$ using $\size(\SUM_n)$ gates; then, compute the resulting bit using at most $O(2^{\log n}/\log n)=o(n)$ gates. For the same reason, any lower bound $\size(f) \ge \alpha$ for
a~symmetric function~$f \in B_n$ implies a~lower bound $\size(\SUM_n) \ge \alpha-o(n)$. Currently, we know the following bounds for $\SUM_n$:
\[2.5n-O(1) \le \size(\SUM_n) \le 4.5n+o(n) \, .\]
The lower bound is~due to~Stockmeyer~\cite{DBLP:journals/mst/Stockmeyer77}, the upper bound is~due to~Demenkov et al.~\cite{DBLP:journals/ipl/DemenkovKKY10}.

A~circuit for $\SUM_n$ can~be constructed from circuits for $\SUM_k$ for some small~$k$. For example,
using full and half adders as~building blocks, one can compute $\SUM_n$ (for any~$n$) by a~circuit of size $5n$ as~follows. Start from $n$~bits $(x_1, \dotsc, x_n)$ of~weight~$0$. While there are three bits of the same weight~$k$, replace them by~two bits of weights~$k$ and~$k+1$ using a~full adder. This way, one gets at~most two bits of~each weight $0,1,\dotsc,l-1$ ($l=\lceil \log_2(n+1)\rceil$) in at most $5(n-l)$ gates (as each full adder reduces the number of~bits). To~leave exactly one bit
of~each weight, it suffices to use at~most~$l$ half
or~full adders ($o(n)$ gates). Let~us denote the
size of the resulting circuit by~$s(n)$. The second row
of~Table~\ref{table:sum} shows the values of~$s(n)$ for
some $n \le 15$ (see~(28) in~\cite{Knuth:2008:ACP:1377542}).

\begin{table}
\begin{center}
\begin{tabular}{lrrrrrrrrrrrrrrrrrrrrrr}
\toprule
$n$ & $2$ & $3$ & $4$ & $5$ & $6$ & $7$ & $8$ & $9$ & $10$ & $15$
\\
\midrule
$s(n)$ &$2$ & $5$ & $9$ & $12$ & $17$ & $20$ & $26$ & $29$& $34$&
$55$
\\
$\size(\SUM_n)$ & $2$ & $5$ & $9$ & $11$
& $\le 16$ & $\le 19$ & $\le 25$ & $\le 27$
& $\le 32$ & $\le 53$
\\
\bottomrule
\end{tabular}
\end{center}
\caption{The first line shows the value of~$n$. The second line gives the size $s(n)$ of~a~circuit for $\SUM_n$
composed out~of half and full adders. The third row shows known bounds for $\size(\SUM_n)$.}
\label{table:sum}
\end{table}

In a~similar fashion, one can get an~upper bound (see Theorem~1 in~\cite{DBLP:conf/date/Kulikov18})
\begin{equation}\label{eq:sumupper}
\size(\SUM_n) \le \frac{\size(\SUM_k)}{k-\lceil \log_2(k+1) \rceil} \cdot n + o(n) \, .
\end{equation}
This motivates the search for efficient circuits
for $\SUM_k$ for small values of~$k$. The bottom row
of~Table~\ref{table:sum} gives upper bounds that
we~were able to~find using the program
(the upper bounds for $n \le 7$ were found
by~Knuth~\cite{Knuth:2008:ACP:1377542}).
The table shows that the first value where $s(n)$ is not
optimal is $n=5$. The best upper bound for $\SUM_n$ given
by~\eqref{eq:sumupper} is $4.75n+o(n)$ for $n=7$. The upper
bound for $n=15$ is $53n/11+o(n)$ which is worse than the
previous upper bound. But if it~turned out that
$\size(\SUM_{15}) \le 52$,
it~would give a~better upper bound.

The found circuits for $\SUM_n$ for $n \le 15$
do~not allow to~improve the strongest known upper bound
$\size(\SUM_n) \le 4.5n+o(n)$ due to~Demenkov et al.~\cite{DBLP:journals/ipl/DemenkovKKY10}. Below,
we~present several interesting observations on~the found circuits.

\subsubsection{Best Known Upper Bound for the SUM Function}
\begin{figure}
\begin{center}
\begin{tikzpicture}
\begin{scope}[scale=.7]
\draw (1,0) rectangle (3,2); \node at (2,1) {$\SUM_3$};
\draw (5,0) rectangle (7,2); \node at (6,1) {$\SUM_3$};
\foreach \n/\x/\y in {3/0/1, 2/1.5/3, 1/2.5/3, 4/5.5/3, 5/6.5/3}
  \node[input] (\n) at (\x,\y) {$x_{\n}$};
\foreach \n/\t/\x/\y in {a1/a_1/2/-1, b1/b_1/6/-1, b0/b_0/8/1}
  \node[input] (\n) at (\x,\y) {$\t$};
\draw[->] (3)--(1,1);
\draw[->] (2)--(1.5,2);
\draw[->] (1)--(2.5,2);
\draw[->] (4)--(5.5,2);
\draw[->] (5)--(6.5,2);
\draw[->] (3,1)--(5,1);
\draw[->] (7,1)--(b0);
\draw[->] (2,0)--(a1);
\draw[->] (6,0)--(b1);

\node at (-1,1) {(a)};
\end{scope}

\begin{scope}[scale=.7,yshift=-60mm]
\draw (1,0) rectangle (7,2); \node at (4,1) {MDFA};
\foreach \n/\x/\y in {3/0/1, 2/1.5/4, 1/2.5/4, 4/5.5/4, 5/6.5/4}
  \node[input] (\n) at (\x,\y) {$x_{\n}$};
\node[gate] (xor1) at (2.5,3) {$\oplus$};
\node[gate] (xor2) at (6.5,3) {$\oplus$};
\foreach \n/\t/\x/\y in {a1/a_1/2/-1, b1/{a_1 \oplus b_1}/6/-1, b0/b_0/8/1}
  \node[input] (\n) at (\x,\y) {$\t$};
\draw[->] (3)--(1,1);
\draw[->] (2)--(1.5,2);
\draw[->] (1) -- (xor1); \draw[->] (2) -- (xor1);
\draw[->] (xor1) -- (2.5,2);
\draw[->] (4)--(5.5,2);
\draw[->] (5)-- (xor2); \draw[->] (xor2) -- (6.5,2); \draw[->] (4) -- (xor2);
\draw[->] (7,1)--(b0);
\draw[->] (2,0)--(a1);
\draw[->] (6,0)--(b1);

\node at (-1,1) {(b)};
\end{scope}

\begin{scope}[yscale=.8, xshift=80mm, yshift=-20mm]
\draw[draw=none, rounded corners=0,fill=gray!20] (0,1.5)--(1,1.5)--(1,2.5)--(2,2.5)--(2,0.5)--(2.5,0.5)--(2.5,-0.5)--(3.5,-0.5)--
(3.5,-2.5)--(0,-2.5)--(0,1.5);

\foreach \n/\x/\y in {1/0/3, 2/1/3, 3/2/3, 4/2.5/1, 5/3.5/1}
  \node[input] (x\n) at (\x, \y) {$x_{\n}$};
\node[gate] (g1) at (0.5,2) {$\oplus$};
\node[gate] (g2) at (1.5,2) {$\oplus$};
\node[gate] (g3) at (0.5,1) {$\lor$};
\node[gate] (g4) at (1.5,1) {$\oplus$};
\node[gate, label=left:$a_1$] (g5) at (0.5,0) {$\oplus$};
\node[gate] (g6) at (2,0) {$\oplus$};
\node[gate] (g7) at (3,0) {$\oplus$};
\node[gate] (g8) at (2,-1) {$>$};
\node[gate, label=right:$b_0$] (g9) at (3,-1) {$\oplus$};
\node[gate, label=right:$a_1 \oplus b_1$] (g10) at (2,-2) {$\oplus$};
\node[gate] (g11) at (1.5,-3) {$>$};

\foreach \f/\t in {x1/g1, x2/g1, x2/g2, x3/g2, g1/g3, g2/g3, g1/g4, g3/g5, g4/g5, x4/g6, g4/g6, x4/g7, x5/g7, g6/g8, g7/g8, g7/g9, g3/g10, g8/g10, g10/g11, g5/g11}
  \draw[->] (\f) -- (\t);

\path (x3) edge[->,bend left] (g4);
\path (g4) edge[->,bend left=20] (g9);

\node at (1.5,-4) {(c)};
\end{scope}
\end{tikzpicture}
\end{center}
\caption{(a)~Two consecutive $\SUM_3$ blocks. (b)~The $\MDFA$ block. (c)~The highlighted part of the optimal circuit
for $\SUM_5$ computes $\MDFA$.}
\label{figure:mdfa}
\end{figure}
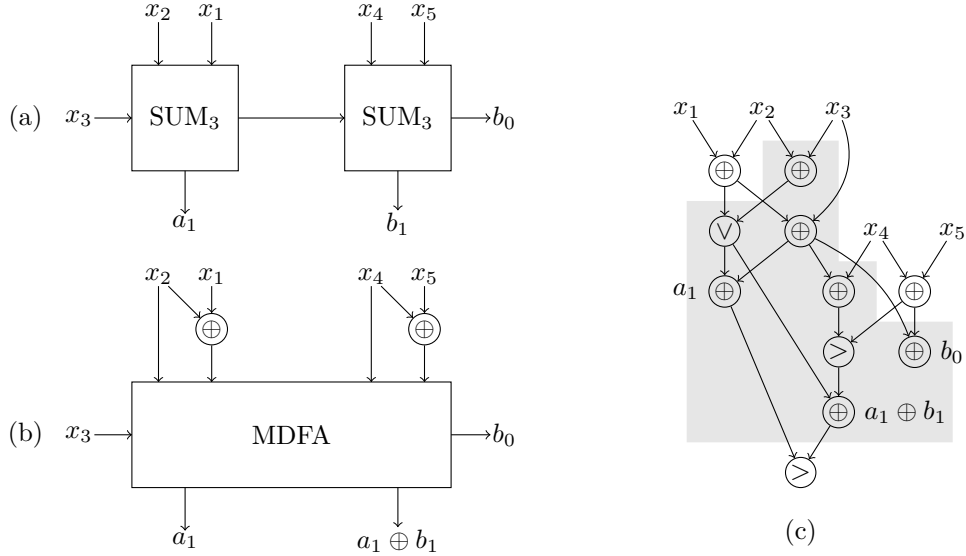

The optimal circuit of size~$11$ for $\SUM_5$ shown in~Figure~\ref{figure:sumfive}(c) can be~used to get an~upper bound $4.5n+o(n)$ for $\size(\SUM_n)$
(though not through~\eqref{eq:sumupper} directly).
To~do this,
consider two consecutive $\SUM_3$ circuits shown in~Figure~\ref{figure:mdfa}(a).
Its specification~is: $x_1+\dotsb+x_5=b_0+2(a_1+b_1)$, its size is~equal to~10. One can construct a~similar block, called MDFA (for modified double full adder), of size~8, whose specification is
\[\MDFA(x_1 \oplus x_2, x_2, x_3, x_4, x_4 \oplus x_5)=(b_0, a_1, a_1 \oplus b_1) \, ,\]
see Figure~\ref{figure:mdfa}(b).

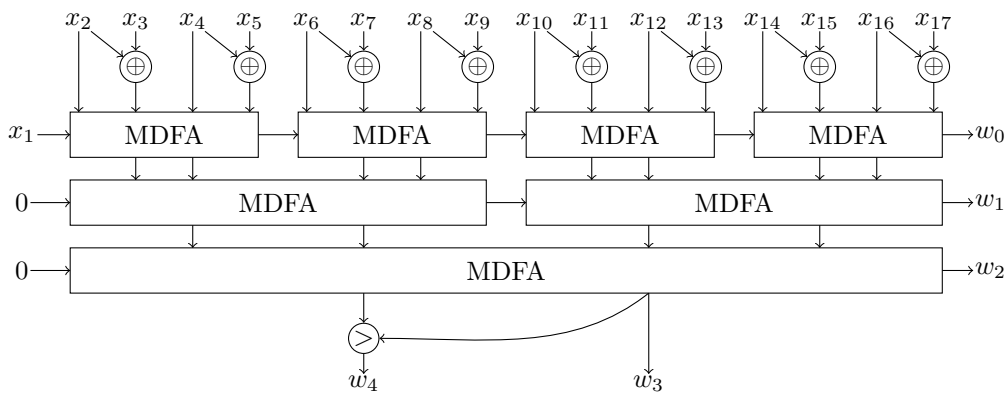
\begin{figure}[!ht]
    \begin{center}
        \begin{tikzpicture}[xscale=0.75, yscale=.6]
            \foreach \n in {2,...,17}
            \node[input] (\n) at (\n,6) {$x_{\n}$};
            \foreach \i in {1,...,8} {
                \tikzmath{\k=int(2*\i); \j=int(2*\i+1);}
                \node[gate] (a) at (\j,5) {$\oplus$};
                \draw[->] (\k) -- (a); \draw[->] (\j) -- (a);
                \draw[->] (a) -- (\j,4); \draw[->] (\k) -- (\k,4);
            }
            \foreach \x/\y/\w in {2/4/3, 6/4/3, 10/4/3, 14/4/3, 2/2.5/7, 10/2.5/7, 2/1/15} {
                \draw (\x-0.15,\y) rectangle (\x+\w+0.15,\y-1);
                \node at (\x+\w/2,\y-0.5) {MDFA};
            }
            \foreach \y/\l/\w in {3.5/x_1/w_0, 2/0/w_1, 0.5/0/w_2} {
                \node[input] (b) at (1,\y) {$\l$};
                \draw[->] (b) -- (1.85,\y);
                \node[input] (c) at (18,\y) {$\w$};
                \draw[->] (17.15,\y) -- (c);
            }
            \foreach \x in {3, 4, 7, 8, 11, 12, 15, 16}
            \draw[->] (\x,3) -- (\x,2.5);
            \foreach \x in {4, 7, 12, 15}
            \draw[->] (\x,1.5) -- (\x,1);
            \foreach \x/\y in {5/3.5, 9/3.5, 13/3.5, 9/2}
            \draw[->] (\x+0.15,\y) -- (\x+0.85,\y);

            \node[input] (w3) at (12,-2) {$w_3$};
            \draw[->] (12,0) -- (w3);
            \node[gate] (x) at (7,-1) {$>$};
            \draw[->] (7,0) -- (x);
            \node[input] (w4) at (7,-2) {$w_4$};
            \draw[->] (x) -- (w4);
            \path (12,0) edge[->,out=-135,in=0] (x);
        \end{tikzpicture}
    \end{center}
    \caption{A~circuit computing $\SUM_{17}$ composed out of~MDFA blocks.}
    \label{figure:sum17}
\end{figure}

The fact that $\MDFA$ uses the encoding $(p, p \oplus q)$ for pairs of bits $(p,q)$, allows to use~it recursively
to~compute $\SUM_n$. As~the original construction is~presented in~\cite{DBLP:journals/ipl/DemenkovKKY10}, below we~give a~sketch
only.
\begin{enumerate}
    \item Compute $x_2 \oplus x_3, x_4 \oplus x_5, \dotsc, x_{n-1} \oplus x_n$ ($n/2$ gates).
    \item Apply at~most~$n/2$ $\MDFA$ blocks (no~more than $4n$~gates).
    \item The last MDFA block outputs two bits: $a$~and~$a\oplus b$. Instead of~them, one needs to~compute $a \oplus b$ and $a \land b$. To~achieve this,
    it~suffices to apply $x>y=(x \land \overline{y})$ operation:
    \[a \land b = a>(a \oplus b)\, .\]
\end{enumerate}
Figure~\ref{figure:sum17} shows an~example for $n=17$.

The $\MDFA$ block was constructed by~Demenkov et al.~\cite{DBLP:journals/ipl/DemenkovKKY10} in~a~semiautomatic
manner. And it turns out that $\MDFA$ is just a~subcircuit of the optimal circuit for $\SUM_5$! See Figure~\ref{figure:mdfa}(c).

\subsubsection{Best Known Circuits for SUM with New Structure}

\begin{figure}[!ht]
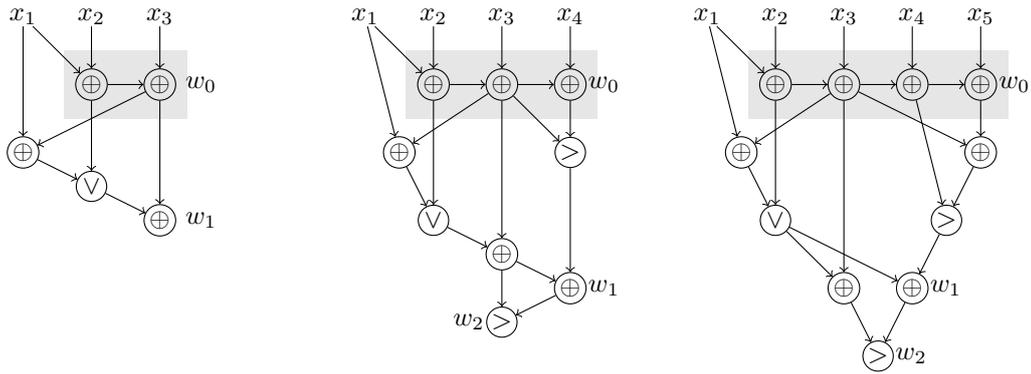

\begin{mypic}
\begin{scope}[scale=.9]
\begin{scope}[xshift=20mm, yshift=30mm]
\draw[draw=none, rounded corners=0,fill=gray!20] (0.6,2.5)--(0.6,3.5)--(2.4,3.5)--(2.4,2.5);

\foreach \n/\x/\y in {1/0/4, 2/1/4, 3/2/4}
  \node[input] (x\n) at (\x, \y) {$x_{\n}$};
\node[gate] (g1) at (1,3) {$\oplus$};
\node[gate,label=right:$w_0$] (g2) at (2,3) {$\oplus$};
\node[gate] (g3) at (0,2) {$\oplus$};
\node[gate] (g4) at (1,1.5) {$\lor$};
\node[gate, label=right:$w_1$] (g5) at (2,1) {$\oplus$};
\foreach \f/\t in {x1/g1, x2/g1, g1/g2, x3/g2, x1/g3, g2/g3, g1/g4, g3/g4, g2/g5, g4/g5}
  \draw[->] (\f) -- (\t);
\end{scope}

\begin{scope}[label distance=-1mm, xshift=70mm, yshift=20mm]
\draw[draw=none, rounded corners=0,fill=gray!20] (0.6,3.5)--(0.6,4.5)--(3.4,4.5)--(3.4,3.5);
\foreach \n/\x/\y in {1/0/5, 2/1/5, 3/2/5, 4/3/5}
  \node[input] (x\n) at (\x, \y) {$x_{\n}$};
\node[gate] (g1) at (1,4) {$\oplus$};
\node[gate] (g2) at (2,4) {$\oplus$};
\node[gate,label=right:$w_0$] (g3) at (3,4) {$\oplus$};

\node[gate] (g4) at (0.5,3) {$\oplus$};
\node[gate] (g5) at (3,3) {$>$};

\node[gate] (g6) at (1,2) {$\lor$};
\node[gate] (g7) at (2,1.5) {$\oplus$};
\node[gate,label=right:$w_1$] (g8) at (3,1) {$\oplus$};

\node[gate,label=left:$w_2$] (g9) at (2,0.5) {$>$};
\foreach \f/\t in {x1/g1, x2/g1, g1/g2, x3/g2, g2/g3, x4/g3, x1/g4, g2/g4, g2/g5, g3/g5, g4/g6, g1/g6, g6/g7, g2/g7, g7/g8, g5/g8, g8/g9, g7/g9}
  \draw[->] (\f) -- (\t);
\end{scope}

\begin{scope}[label distance=-1mm, xshift=120mm, yshift=20mm]
\draw[draw=none, rounded corners=0,fill=gray!20] (0.6,3.5)--(0.6,4.5)--(4.4,4.5)--(4.4,3.5);
\foreach \n/\x/\y in {1/0/5, 2/1/5, 3/2/5, 4/3/5, 5/4/5}
  \node[input] (x\n) at (\x, \y) {$x_{\n}$};
\node[gate] (g1) at (1,4) {$\oplus$};
\node[gate] (g2) at (2,4) {$\oplus$};
\node[gate] (g3) at (3,4) {$\oplus$};
\node[gate,label=right:$w_0$] (g4) at (4,4) {$\oplus$};

\node[gate] (g5) at (0.5,3) {$\oplus$};
\node[gate] (g6) at (4,3) {$\oplus$};

\node[gate] (g7) at (1,2) {$\lor$};
\node[gate] (g8) at (3.5,2) {$>$};

\node[gate] (g9) at (2,1) {$\oplus$};
\node[gate,label=right:$w_1$] (g10) at (3,1) {$\oplus$};

\node[gate,label=right:$w_2$] (g11) at (2.5,0) {$>$};

\foreach \f/\t in {x1/g1, x2/g1, g1/g2, x3/g2, g2/g3, x4/g3, g3/g4, x5/g4, x1/g5, g2/g5, g2/g6, g4/g6, g1/g7, g5/g7, g3/g8, g6/g8, g2/g9, g7/g9, g8/g10, g7/g10, g9/g11, g10/g11}
  \draw[->] (\f) -- (\t);
\end{scope}

\end{scope}
\end{mypic}
\caption{Optimal circuits computing $\SUM_n$
for $n=3,4,5$ with a~specific structure: every input,
except for~$x_1$, has out-degree~one.}
\label{figure:xorsum}
\end{figure}

For many upper bounds from the bottom row of~Table~\ref{table:sum}, we found circuits with the following interesting structure: the first thing the circuit computes is $x_1 \oplus x_2 \oplus \dotsb \oplus x_n$; moreover the variables $x_2, \dotsc, x_n$ are used for this only. This is best illustrated by an~example~--- see Figure~\ref{figure:xorsum}.

These circuits can be~found using the following code.
It~demonstrates two new useful features: fixing gates and forbidding wires between some pairs of~gates.

\inputminted[firstline=71,lastline=86]{python}{../tutorial.py}

\subsubsection{Optimal Circuits for Counting Modulo~4}
The optimal circuit for $\SUM_5$ can be~used to~construct
an~optimal circuit of~size $2.5n+O(1)$ for $\MOD_{4,r}^n$ due
to~Stockmeyer~\cite{DBLP:journals/mst/Stockmeyer77}
(recall that $\MOD_n^{4,r}(x_1, \dotsc, x_n)=[x_1+\dotsb+x_n \equiv r \bmod 4]$).
To~do this, note that
there is a~subcircuit (of the circuit at~Figure~\ref{figure:sumfive}(c)) of size~9 that computes the two least significant bits ($w_0,w_1$) of $x_1+\dotsb+x_5$ (one removes the gates $g_5, w_2$). To~compute $x_1+\dotsb+x_n \bmod 4$, one first applies $\frac n4$ such blocks and then computes the parity of~the resulting bits of weight~$1$
(every block takes four fresh inputs as~well as~one bit of~weight~$0$ from the previous block).
The total size is~$9 \cdot \frac n4 + \frac n4=2.5n$.
Thus,
the circuit that Stockmeyer constructed in~1977 by~hand,
nowadays can be~found automatically in~a~few seconds.

\subsection{Modulo-3 Function}
In~\cite{DBLP:conf/sat/KojevnikovKY09}, Kojevnikov et~al.
presented circuits of~size $3n+O(1)$ for $\MOD_n^{3,r}$ (for any~$r$). Later,
Knuth~\cite[solution to exercise~$480$]{Knuth:2015:ACP:2898950} analyzed their construction and proved an~upper
bound $3n-4$. Also, by~finding the exact values
for $\size(\MOD_n^{3,r})$ for all $3 \le n \le 5$ and all $0 \le r \le 2$, he~made the conjecture~\eqref{conjecture}.
Using our program, we~proved the conjectured upper bound for all~$n$.

\begin{theorem}\label{theorem:mod3upper}
For all $n \ge 3$ and all $r \in \{0,1,2\}$,
\[\size(\MOD_n^{3,r}) \le 3n-5-[(n+r) \equiv 0\bmod 3] \,. \]
\end{theorem}

To~prove Knuth's conjecture, one also needs to~prove a~lower bound on $\size(\MOD_n^{3,r})$. The currently strongest known lower bound for $\size(\MOD_n^{3,r})$
is~$2.5n-O(1)$ due to~Stockmeyer~\cite{DBLP:journals/mst/Stockmeyer77}
(and no~stronger lower bound is known for any other symmetric function).

\begin{proof}
As~in~\cite{DBLP:conf/sat/KojevnikovKY09},
we~construct the required circuit out of~constant size blocks. Schematically, the circuit looks as~follows.

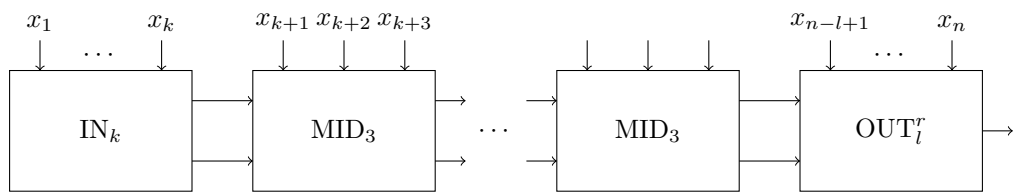
\begin{center}
\begin{tikzpicture}[scale=.8]

\foreach \x/\k in {0.5/1, 2.5/k, 4.5/{k+1}, 5.5/{k+2}, 6.5/{k+3}, 13.5/{n-l+1}, 15.5/n}
  \node[above] at (\x,2.5) {$x_{\k}$};

\foreach \x in {0, 4, 9, 13}
  \draw (\x,0) rectangle (\x+3,2);

\foreach \x in {0.5, 2.5, 4.5, 5.5, 6.5, 9.5, 10.5, 11.5, 13.5, 15.5}
  \draw[->] (\x,2.5) -- (\x,2);

\foreach \x/\y in {1.5/2.25, 14.5/2.25, 8/1}
  \node at (\x,\y) {$\dotsb$};

\foreach \a/\b in {3/4, 12/13, 7/7.5, 8.5/9} {
  \draw[->] (\a,1.5) -- (\b,1.5);
  \draw[->] (\a,0.5) -- (\b,0.5);
}

\draw[->] (16,1) -- (16.5,1);
\node at (1.5, 1) {$\IN_k$};
\node at (5.5, 1) {$\MID_3$};
\node at (10.5, 1) {$\MID_3$};
\node at (14.5, 1) {$\OUT_l^r$};
\end{tikzpicture}
\end{center}
Here, the $n$~input bits are passed from above.
What is passed from block to block (from left to~right)
is the pair of bits $(r_0, r_1)$ encoding the current remainder~$r$ modulo~$3$ as~follows: if $r=0$, then $(r_0,r_1)=(0,0)$; if $r=1$, then $(r_0,r_1)=(0,1)$; if $r=2$, then $r_0=1$. The first block $\IN_k$ takes the first $k$~input bits and computes the remainder of their sum modulo~$3$. It is followed by a~number of~$\MID_3$ blocks each of which takes the current remainder and three new input bits and computes the new remainder. Finally, the block~$\OUT_l^r$ takes the remainder and the last~$l$ input bits and outputs $\MOD_n^{3,r}$. The integers $k,l$ take values in~$\{2,3,4\}$ and $\{1,2,3\}$, respectively. Their exact values depend on~$r$ and $n \bmod 3$ as~described below.

The theorem follows from the following upper bounds
on~the circuit size of the just introduced functions:
$\size(\IN_2) \le 2$,
$\size(\IN_3) \le 5$,
$\size(\IN_4) \le 7$,
$\size(\MID_3) \le 9$,
$\size(\OUT_2^0) \le 5$,
$\size(\OUT_1^1) \le 2$,
$\size(\OUT_3^2) \le 8$.
The corresponding circuits are presented in the Appendix
by a~straightforward \texttt{Python} code that verifies their correctness. (The presented code proves the mentioned upper bounds by~providing explicit circuits. We~have also verified that no~smaller circuits exist meaning that the
inequalities above are in fact equalities.)
Table~\ref{table:parameters} shows how to~combine
the blocks to~get a~circuit computing~$\MOD_n^{3,r}$ of the required size.

\begin{table}[!ht]
\begin{center}
\begin{tabular}{cccc}
\toprule
& $n=3t$ & $n=3t+1$ & $n=3t+2$\\
\midrule
$r=0$
& $(4, t-2, 2)$, $(7, 5)$, $3n-6$
& $(2, t-1, 2)$, $(2, 5)$, $3n-5$
& $(3, t-1, 2)$, $(5, 5)$, $3n-5$
\\
$r=1$
& $(2, t-1, 1)$, $(2, 2)$, $3n-5$
& $(3, t-1, 1)$, $(5, 2)$, $3n-5$
& $(4, t-1, 1)$, $(7, 2)$, $3n-6$
\\
$r=2$
& $(3, t-2, 3)$, $(5, 8)$, $3n-5$
& $(4, t-2, 3)$, $(7, 8)$, $3n-6$
& $(2, t-1, 3)$, $(2, 8)$, $3n-5$
\\
\bottomrule
\end{tabular}
\end{center}
\caption{Choosing parameters $k, m, l$ depending on $n \bmod 3$ and~$r$. The circuit is composed out of~blocks
as~follows: $\IN_k + m \times \MID_3 + \OUT_r^l$. For each
pair $(n \bmod 3, r)$ we show three things: the triple
$(k, m, l)$; the sizes of~two blocks: $\size(\IN_k)$ and $\size(\OUT_r^l)$; the size of~the resulting circuit
computed as $s=\size(\IN_k)+9m+\size(\OUT_r^l)$. For example, the top left cell is read as follows: when $r=0$ and $n=3t$, we set $k=4,m=t-2,l=2$; the resulting circuit
is~then $\IN_4 + (t-2) \times \MID_3 + \OUT_0^2$; since $\size(\IN_4)=7$ and $\size(\OUT_0^2)=5$, the size of~the circuit is $7+9(t-2)+5=9t-6=3n-6$.
There are three corners cases that are not well-defined as~they require the
number of~$\MID$ blocks to~be negative ($k=t-2$): $(n=3,r=0)$, $(n=3,r=2)$, and $(n=4, r=2)$. The corresponding circuits are
given in the~Appendix.}
\label{table:parameters}
\end{table}
\end{proof}

\subsection{Threshold Function}
\begin{theorem}
    For any $4 \le k = O(1)$,
    \[
        \size(\THR_n^k) \le (4.5-2^{2 - \lceil \log_2k\rceil})n+o(n) \, .
    \]
\end{theorem}
\begin{proof}
    For a~sequence of~$2m$ formal variables $y_1, z_1, \dotsc, y_m, z_m$,
    consider a~function $g \in B_{2m}$ that takes
    \[y_1, y_1 \oplus z_1, y_2, y_2 \oplus z_2, \dotsc, y_m, y_m \oplus z_m\]
    as~input and outputs $\THR_{2m}^2(y_1, z_1, \dotsc, y_m, z_m)$.
    Note that $\THR_{2m}^2(y_1, z_1, \dotsc, y_m, z_m)=1$ iff there is
    a~pair containing two~$1$'s or there are two pairs each containing at~least one~$1$: $\THR_{2m}^2(y_1, z_1, \dotsc, y_m, z_m)=1$ iff there exists
    $1 \le i \le m$ such that $y_i=z_i=1$ or $\THR_m^2(y_1 \oplus z_1, \dotsc, y_m \oplus z_m)=1$. The condition $y_i=z_i=1$ can be~computed
    through $y_i$ and $y_i \oplus z_i$ using a~single binary gate:
    \[(y_i \land z_i)=(y_i \land \overline{(y_i \oplus z_i)}) \, .\]
    Thus,
    \[g(y_1, y_1 \oplus z_1,\dotsc, y_m, y_m \oplus z_m)=\THR_m^2(y_1 \oplus z_1, \dotsc, y_m \oplus z_m) \lor \bigvee_{i=1}^{m}(y_i \land \overline{(y_i \oplus z_i)}) \, .\]
    Now, $\size(\THR_m^2) \le 2m+o(m)$ as~shown by~Dunne~\cite{Dunne84}.
    Also, clearly,
    \[\size\left(\bigvee_{i=1}^{m}(y_i \land \overline{(y_i \oplus z_i)}) \right)\le 2m-1 \, .\]
    Thus,
    \begin{equation}\label{eq:g}
        \size(g) \le 4m+o(m) \, .
    \end{equation}

    First, consider the case $k=2^t$ where $t \ge 2$ is an~integer.
    To~construct a~circuit for $\THR_n^k$, we~first apply $t-1$
    layers of~$\MDFA$'s (as~in Figure~\ref{figure:sum17}). It~takes
    \[\frac{n}{2} + n\sum\limits_{i=1}^{t-1} 2^{2-i}=(4.5 - 2^{3-t})n\] gates.
    As a~result, we get bits \[w_0, \dotsc, w_{t-2}, a_1, a_1 \oplus b_1, \dotsc, a_m, a_m \oplus b_m \,, \]
    where $m=n/2^t$,
    such that
    \[x_1+\dotsb+x_n=w_0+2w_1+\dotsb+2^{t-2}w_{t-2}+2^{t-1}(a_1+b_1+\dotsb+a_m+b_m) \, .\]
    Note that $w_0+2w_1+\dotsb+2^{t-2}w_{t-2} < 2^{t-1}$. Hence,
    \[[x_1+\dotsb+x_n \ge 2^t]=[a_1+b_1+\dotsb+a_m+b_m \ge 2]\,.\]
    Thus, it~remains to~compute the function~$g$
    given $2m$~bits $a_1, a_1 \oplus b_1, \dotsc, a_m, a_m \oplus b_m$.
    By~\eqref{eq:g}, it takes $4m+o(m)=2^{2-t}n+o(n)$ gates.
    The total size of the constructed circuit~is
    \[(4.5-2^{3-t}+2^{2-t})n+o(n)=(4.5-2^{2-t})n+o(n) \, .\]

    Now, assume that $2^{t-1}<k<2^t$. Then, $\lceil \log_2k \rceil = t$
    and $2^t<2k=O(1)$. Clearly,
    \[[x_1+\dotsb+x_n \ge k]=[(2^t-k)+x_1+\dotsb+x_n \ge 2^t] \, .\]
    By~the previous argument, there exists a~circuit~$C$ computing $\THR_{n+(2^t-k)}^{2^t}$ of~size
    \[(4.5-2^{2-t})(n+(2^t-k))+o(n)=(4.5-2^{2-t})n+o(n) \, .\]
    By~replacing arbitrary $(2^t-k)$ inputs of~$C$ by~$1$'s,
    one gets a~circuit of~the required size computing~$\THR_n^k$.
\end{proof}

\begin{figure}[t]
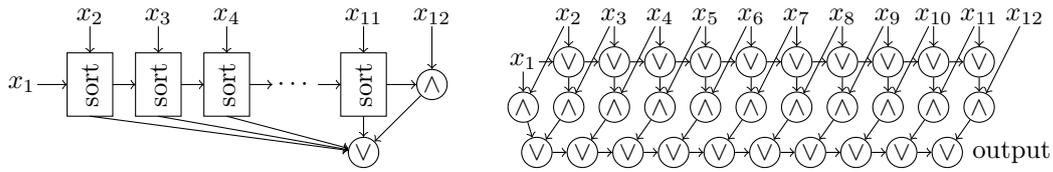

\begin{mypic}
\begin{scope}[scale=.6]
\begin{scope}[xshift=20mm, yshift=30mm]

\foreach \n/\x/\y in {1/0/11, 2/1/12, 3/2/12, 4/3/12, 5/4/12, 6/5/12, 7/6/12, 8/7/12, 9/8/12, 10/9/12, 11/10/12, 12/11/12}
  \node[input] (x\n) at (\x, \y) {$x_{\n}$};

\foreach \n/\x/\y in {2/1/11, 4/2/11, 6/3/11, 8/4/11, 10/5/11, 12/6/11, 14/7/11, 16/8/11, 18/9/11, 20/10/11}
  \node[gate] (g\n) at (\x, \y) {$\lor$};

\foreach \n/\x/\y in {1/0/10, 3/1/10, 5/2/10, 7/3/10, 9/4/10, 11/5/10, 13/6/10, 15/7/10, 17/8/10, 19/9/10}
  \node[gate] (g\n) at (\x, \y) {$\land$};

\foreach \n/\x/\y in {21/0.3/9, 22/1.3/9, 23/2.3/9, 24/3.3/9, 25/4.3/9, 26/5.3/9, 27/6.3/9, 28/7.3/9, 29/8.3/9}
  \node[gate] (g\n) at (\x, \y) {$\lor$};

\node[gate] (g30) at (10,10) {$\land$};
\node[gate, label=right:output] (g31) at (9.3,9) {$\lor$};

\foreach \f/\t in {x1/g1, x1/g2, x2/g1, x2/g2,
                   g2/g3, g2/g4, x3/g3, x3/g4,
                   g4/g5, g4/g6, x4/g5, x4/g6,
                   g6/g7, g6/g8, x5/g7, x5/g8,
                   g8/g9, g8/g10, x6/g9, x6/g10,
                   g10/g11, g10/g12, x7/g11, x7/g12,
                   g12/g13, g12/g14, x8/g13, x8/g14,
                   g14/g15, g14/g16, x9/g15, x9/g16,
                   g16/g17, g16/g18, x10/g17, x10/g18,
                   g18/g19, g18/g20, x11/g19, x11/g20,
                   g1/g21, g3/g21, g21/g22, g5/g22,
                   g22/g23, g7/g23, g23/g24, g9/g24,
                   g24/g25, g11/g25, g25/g26, g13/g26,
                   g26/g27, g15/g27, g27/g28, g17/g28,
                   g28/g29, g19/g29,
                   x12/g30, g20/g30, g29/g31, g30/g31}
  \draw[->] (\f) -- (\t);
\end{scope}

\end{scope}

\begin{scope}[scale=.6,xshift=-90mm,yshift=100mm]

\node[input] (x1) at (0, 3.5) {$x_{1}$};
\foreach \n/\x/\y in {2/1.5/5, 3/3/5, 4/4.5/5, 11/7.5/5, 12/9/5}
  \node[input] (x\n) at (\x, \y) {$x_{\n}$};

\node at (6,3.5) {$\dotsb$};
\foreach \x/\y/\z in {1/2/1.5, 2.5/3.5/3, 4/5/4.5, 7/8/7.5} {
  \draw (\x,2.8) rectangle (\y,4.2);
  \node[rotate=90] at (\z,3.5) {sort};
}

\node[gate] (g1) at (9,3.5) {$\land$};
\node[gate] (g2) at (7.5,2) {$\lor$};

\draw[->] (x1) -- (1,3.5);
\draw[->] (x2) -- (1.5,4.2);
\draw[->] (x3) -- (3,4.2);
\draw[->] (x4) -- (4.5,4.2);
\draw[->] (x11) -- (7.5,4.2);
\draw[->] (x12) -- (g1);

\draw[->] (2,3.5) -- (2.5,3.5);
\draw[->] (3.5,3.5) -- (4,3.5);
\draw[->] (5,3.5) -- (5.5,3.5);
\draw[->] (6.5,3.5) -- (7,3.5);
\draw[->] (8,3.5) -- (g1);

\draw[->] (1.5,2.8) -- (g2);
\draw[->] (3,2.8) -- (g2);
\draw[->] (4.5,2.8) -- (g2);
\draw[->] (7.5,2.8) -- (g2);
\draw[->] (g1) -- (g2);

\foreach \f/\t in {}
  \draw[->] (\f) -- (\t);
\end{scope}
\end{mypic}

\caption{A~circuit of~size~$31$ for~$\THR_{12}^2$:
(a)~block structure and (b)~gate structure.
The $\operatorname{SORT}(u,v)$ block sorts two input bits
as~follows: $\operatorname{SORT}(u,v)=(\min\{u,v\}, \max\{u,v\})=(u \land v, u \lor v)$. The circuit performs
one and a~half iterations of~the bubble sort algorithm:
one first finds the maximum bit among $n$~input bits; then,
it~remains to~compute the disjunction of the remaining $n-1$ bits to~check whether there~is at~least one~$1$ among them.
In~general, this leads to~a~circuit of~size $3n-5$.}
\label{figure:thr31}
\end{figure}

\begin{figure}[t]
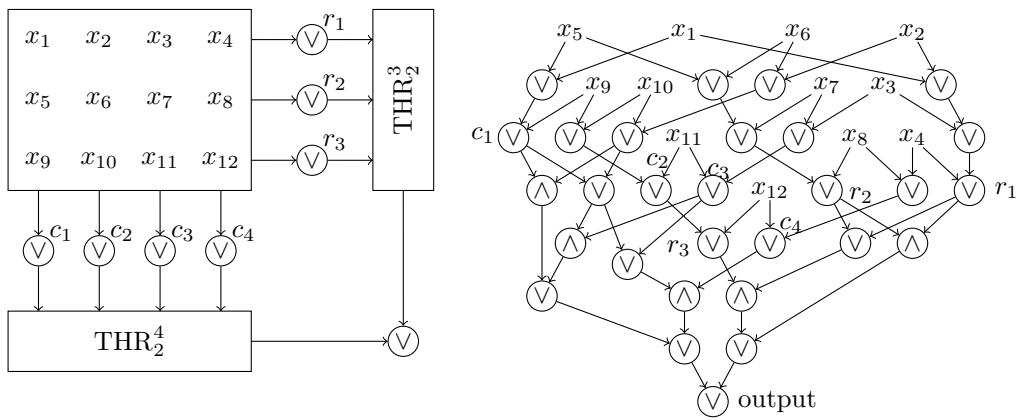

\begin{mypic}
\begin{scope}[xscale=.75,yscale=.7]
\begin{scope}[xshift=20mm, yshift=40mm]

\node[input] (x5) at (2, 7) {$x_5$};
\node[input] (x1) at (4, 7) {$x_1$};
\node[input] (x6) at (6, 7) {$x_6$};
\node[input] (x2) at (8, 7) {$x_2$};

\node[gate] (g9) at (1.5,6) {$\lor$};
\node[input] (x9) at (2.5, 6) {$x_9$};
\node[input] (x10) at (3.5, 6) {$x_{10}$};
\node[gate] (g3) at (4.5,6) {$\lor$};
\node[gate] (g11) at (5.5,6) {$\lor$};
\node[input] (x7) at (6.5, 6) {$x_7$};
\node[input] (x3) at (7.5, 6) {$x_3$};
\node[gate] (g0) at (8.5,6) {$\lor$};

\node[gate, label={[shift={(-0.4,-0.4)}]$c_1$}] (g10) at (1,5) {$\lor$};
\node[gate] (g6) at (2,5) {$\lor$};
\node[gate, label={[shift={(0.4,-0.74)}]$c_2$}] (g12) at (3,5) {$\lor$};
\node[input] (x11) at (4, 5) {$x_{11}$};
\node[gate] (g4) at (5,5) {$\lor$};
\node[gate] (g13) at (6,5) {$\lor$};
\node[input] (x8) at (7, 5) {$x_8$};
\node[input] (x4) at (8, 5) {$x_4$};
\node[gate] (g1) at (9,5) {$\lor$};

\node[gate] (g22) at (1.5,4) {$\land$};
\node[gate] (g21) at (2.5,4) {$\lor$};
\node[gate] (g7) at (3.5,4) {$\lor$};
\node[gate, label={[shift={(0.08,-0.18)}]$c_3$}] (g14) at (4.5,4) {$\lor$};
\node[input] (x12) at (5.5, 4) {$x_{12}$};
\node[gate, label={[shift={(0.45,-0.50)}]$r_2$}] (g5) at (6.5,4) {$\lor$};
\node[gate] (g15) at (8,4) {$\lor$};
\node[gate, label=right:$r_1$] (g2) at (9,4) {$\lor$};

\node[gate] (g24) at (2,3) {$\land$};
\node[gate] (g23) at (3,2.6) {$\lor$};
\node[gate, label={[shift={(-0.45,-0.5)}]$r_3$}] (g8) at (4.5,3) {$\lor$};
\node[gate, label={[shift={(0.28,-0.2)}]$c_4$}] (g16) at (5.5,3) {$\lor$};
\node[gate] (g17) at (7,3) {$\lor$};
\node[gate] (g18) at (8,3) {$\land$};

\node[gate] (g25) at (1.5,2) {$\lor$};
\node[gate] (g26) at (4,2) {$\land$};
\node[gate] (g19) at (5,2) {$\land$};

\node[gate] (g27) at (4,1) {$\lor$};
\node[gate] (g20) at (5,1) {$\lor$};

\node[gate, label=right:output] (g28) at (4.5,0) {$\lor$};

\foreach \f/\t in {x5/g9, x1/g9, x5/g3, x6/g3, x6/g11, x2/g11, x2/g0, x1/g0,
                   g9/g10, x9/g10, x9/g6, x10/g6, x10/g12, g11/g12, g3/g4, x7/g4, x7/g13, x3/g13, x3/g1, g0/g1,
                   g10/g22, g10/g21, g6/g7, g12/g22, g12/g21, x11/g7, x11/g14, g4/g5, g13/g14, x8/g5, x8/g15, x4/g15, x4/g2, g1/g2,
                   g22/g25, g21/g24, g21/g23, g7/g8, g14/g24, g14/g23, x12/g8, x12/g16, g5/g17, g5/g18, g15/g16, g2/g17, g2/g18,
                   g24/g25, g23/g26, g8/g19, g16/g26, g17/g19, g18/g20,
                   g25/g27, g26/g27, g19/g20,
                   g27/g28, g20/g28}
  \draw[->] (\f) -- (\t);
\end{scope}

\end{scope}

\begin{scope}[scale=.8,xshift=-60mm,yshift=25mm]
\draw (0.5,4.5) rectangle (4.5,7.5); \node at (0,0) {};
\foreach \n/\x/\y in {1/1/7, 2/2/7, 3/3/7, 4/4/7, 5/1/6, 6/2/6, 7/3/6, 8/4/6, 9/1/5, 10/2/5, 11/3/5, 12/4/5}
  \node[input] (x\n) at (\x,\y) {$x_{\n}$};

\foreach \n/\x/\y in {1/5.5/7, 2/5.5/6, 3/5.5/5}
  \node[gate, label={[shift={(0.3,-0.2)}]$r_\n$}] (r\n) at (\x, \y) {$\lor$};
\draw[->] (4.5,7)--(r1);
\draw[->] (4.5,6)--(r2);
\draw[->] (4.5,5)--(r3);

\foreach \n/\x/\y in {1/1/3.5, 2/2/3.5, 3/3/3.5, 4/4/3.5}
  \node[gate, label={[shift={(0.3,-0.2)}]$c_\n$}] (c\n) at (\x, \y) {$\lor$};
\draw[->] (1,4.5)--(c1);
\draw[->] (2,4.5)--(c2);
\draw[->] (3,4.5)--(c3);
\draw[->] (4,4.5)--(c4);

\draw (0.5,1.5) rectangle (4.5,2.5);
\node at (2.5,2) {$\THR_2^4$};
\draw[->] (c1) -- (1,2.5);
\draw[->] (c2) -- (2,2.5);
\draw[->] (c3) -- (3,2.5);
\draw[->] (c4) -- (4,2.5);

\draw (6.5,4.5) rectangle (7.5,7.5);
\node[rotate=90] at (7,6) {$\THR_2^3$};
\draw[->] (r1) -- (6.5,7);
\draw[->] (r2) -- (6.5,6);
\draw[->] (r3) -- (6.5,5);

\node[gate] (g) at (7, 2) {$\lor$};
\draw[->] (4.5,2)--(g);
\draw[->] (7,4.5)--(g);

\end{scope}

\end{mypic}
\caption{A~circuit of~size~$29$ for $\THR_{12}^2$: (a)~block structure and (b)~gate structure. It~implements a~clever trick by~Dunne~\cite{Dunne84}. Organize $12$~input bits
into a~$3 \times 4$ table. Compute disjunctions $r_1,r_2,r_3$ of~the rows and disjunctions $c_1,c_2,c_3,c_4$ of~the columns. Then, there are at least two~$1$'s among
$x_1, \dotsc, x_{12}$ if~and only if~there are at~least
two~$1$'s among either $r_1,r_2,r_3$ or $c_1,c_2,c_3,c_4$.
This allows to~proceed recursively. In~general, it~leads
to~a~circuit of~size $2n+o(n)$. (Sergeev~\cite{Sergeev2020} showed recently that the \emph{monotone} circuit size of $\THR_n^2$ is $2n+\Theta(\sqrt n)$.)}
\label{figure:thr29}
\end{figure}

We~conclude by~presenting an~example of a~reasonably small circuit
that our program fails to~improve though a~better circuit
is~known.
Figures~\ref{figure:thr31} and~\ref{figure:thr29} show circuits of~size~$31$ and~$29$ for $\THR_{12}^2$.
They are
quite different and our program is~not able to~find out
that the circuit
of~size~$31$ is~suboptimal.
One can construct the two circuits in~the program as~follows.

\inputminted[firstline=90,lastline=96]{python}{../tutorial.py}

\section{Further Directions}
In~the paper, we~focus mainly on~proving asymptotic upper bounds for function families (that is, upper bounds that hold for every input size).
A~natural further step is~to~apply the program
to~specific circuits that are used in~practice.
It~would also be~interesting to~extend the program
so~that it~is able to~discover the circuit from Figure~\ref{figure:thr29}.

\bibliography{circuits}

\begin{thebibliography}{10}

\bibitem{abc}
\url{https://github.com/berkeley-abc/abc}.

\bibitem{DBLP:conf/cade/BrakensiekHMN20}
Joshua Brakensiek, Marijn Heule, John Mackey, and David Narv{\'{a}}ez.
\newblock The resolution of {K}eller's conjecture.
\newblock In Nicolas Peltier and Viorica Sofronie{-}Stokkermans, editors, {\em
  Automated Reasoning - 10th International Joint Conference, {IJCAR} 2020,
  Paris, France, July 1-4, 2020, Proceedings, Part {I}}, volume 12166 of {\em
  Lecture Notes in Computer Science}, pages 48--65. Springer, 2020.
\newblock \href {https://doi.org/10.1007/978-3-030-51074-9\_4}
  {\path{doi:10.1007/978-3-030-51074-9\_4}}.

\bibitem{git-improvement}
\url{https://github.com/alexanderskulikov/circuit_improvement}.

\bibitem{DBLP:journals/ipl/DemenkovKKY10}
Evgeny Demenkov, Arist Kojevnikov, Alexander~S. Kulikov, and Grigory
  Yaroslavtsev.
\newblock New upper bounds on the boolean circuit complexity of symmetric
  functions.
\newblock {\em Inf. Process. Lett.}, 110(7):264--267, 2010.
\newblock \href {https://doi.org/10.1016/j.ipl.2010.01.007}
  {\path{doi:10.1016/j.ipl.2010.01.007}}.

\bibitem{Dunne84}
Paul~E. Dunne.
\newblock {\em Techniques for the analysis of monotone Boolean networks}.
\newblock PhD thesis, University of Warwick, 1984.

\bibitem{DBLP:conf/sat/FichteLS17}
Johannes~Klaus Fichte, Neha Lodha, and Stefan Szeider.
\newblock Sat-based local improvement for finding tree decompositions of small
  width.
\newblock In Serge Gaspers and Toby Walsh, editors, {\em Theory and
  Applications of Satisfiability Testing - {SAT} 2017 - 20th International
  Conference, Melbourne, VIC, Australia, August 28 - September 1, 2017,
  Proceedings}, volume 10491 of {\em Lecture Notes in Computer Science}, pages
  401--411. Springer, 2017.
\newblock \href {https://doi.org/10.1007/978-3-319-66263-3\_25}
  {\path{doi:10.1007/978-3-319-66263-3\_25}}.

\bibitem{DBLP:conf/focs/FindGHK16}
Magnus~Gausdal Find, Alexander Golovnev, Edward~A. Hirsch, and Alexander~S.
  Kulikov.
\newblock A better-than-3n lower bound for the circuit complexity of an
  explicit function.
\newblock In Irit Dinur, editor, {\em {IEEE} 57th Annual Symposium on
  Foundations of Computer Science, {FOCS} 2016, 9-11 October 2016, Hyatt
  Regency, New Brunswick, New Jersey, {USA}}, pages 89--98. {IEEE} Computer
  Society, 2016.
\newblock \href {https://doi.org/10.1109/FOCS.2016.19}
  {\path{doi:10.1109/FOCS.2016.19}}.

\bibitem{DBLP:conf/sat/IgnatievMM18}
Alexey Ignatiev, Ant{\'{o}}nio Morgado, and Jo{\~{a}}o Marques{-}Silva.
\newblock {PySAT}: {A} {Python} toolkit for prototyping with {SAT} oracles.
\newblock In Olaf Beyersdorff and Christoph~M. Wintersteiger, editors, {\em
  Theory and Applications of Satisfiability Testing - {SAT} 2018 - 21st
  International Conference, {SAT} 2018, Held as Part of the Federated Logic
  Conference, FloC 2018, Oxford, UK, July 9-12, 2018, Proceedings}, volume
  10929 of {\em Lecture Notes in Computer Science}, pages 428--437. Springer,
  2018.
\newblock \href {https://doi.org/10.1007/978-3-319-94144-8\_26}
  {\path{doi:10.1007/978-3-319-94144-8\_26}}.

\bibitem{knuthreduction}
\url{http://www-cs-faculty.stanford.edu/~knuth/programs.html}.

\bibitem{Knuth:2008:ACP:1377542}
Donald~E. Knuth.
\newblock {\em The Art of Computer Programming, Volume 4, Fascicle 0:
  Introduction to Combinatorial Algorithms and Boolean Functions (Art of
  Computer Programming)}.
\newblock Addison-Wesley Professional, 1 edition, 2008.

\bibitem{Knuth:2015:ACP:2898950}
Donald~E. Knuth.
\newblock {\em The Art of Computer Programming, Volume~4, Fascicle~6:
  Satisfiability}.
\newblock Addison-Wesley Professional, 1st edition, 2015.

\bibitem{DBLP:conf/sat/KojevnikovKY09}
Arist Kojevnikov, Alexander~S. Kulikov, and Grigory Yaroslavtsev.
\newblock Finding efficient circuits using {SAT}-solvers.
\newblock In Oliver Kullmann, editor, {\em Theory and Applications of
  Satisfiability Testing - {SAT} 2009, 12th International Conference, {SAT}
  2009, Swansea, UK, June 30 - July 3, 2009. Proceedings}, volume 5584 of {\em
  Lecture Notes in Computer Science}, pages 32--44. Springer, 2009.
\newblock \href {https://doi.org/10.1007/978-3-642-02777-2_5}
  {\path{doi:10.1007/978-3-642-02777-2_5}}.

\bibitem{DBLP:conf/date/Kulikov18}
Alexander~S. Kulikov.
\newblock Improving circuit size upper bounds using sat-solvers.
\newblock In Jan Madsen and Ayse~K. Coskun, editors, {\em 2018 Design,
  Automation {\&} Test in Europe Conference {\&} Exhibition, {DATE} 2018,
  Dresden, Germany, March 19-23, 2018}, pages 305--308. {IEEE}, 2018.
\newblock \href {https://doi.org/10.23919/DATE.2018.8342026}
  {\path{doi:10.23919/DATE.2018.8342026}}.

\bibitem{DBLP:journals/eccc/KulikovS21}
Alexander~S. Kulikov and Nikita Slezkin.
\newblock Sat-based circuit local improvement.
\newblock {\em Electron. Colloquium Comput. Complex.}, 28:44, 2021.
\newblock URL: \url{https://eccc.weizmann.ac.il/report/2021/044}.

\bibitem{DBLP:journals/tocl/LodhaOS19}
Neha Lodha, Sebastian Ordyniak, and Stefan Szeider.
\newblock A {SAT} approach to branchwidth.
\newblock {\em {ACM} Trans. Comput. Log.}, 20(3):15:1--15:24, 2019.
\newblock \href {https://doi.org/10.1145/3326159} {\path{doi:10.1145/3326159}}.

\bibitem{Lup59}
Oleg Lupanov.
\newblock {A method of circuit synthesis}.
\newblock {\em Izvestiya VUZov, Radiofizika}, 1:120--140, 1959.

\bibitem{M56}
David~E. Muller.
\newblock Complexity in electronic switching circuits.
\newblock {\em IRE Transactions on Electronic Computers}, EC-5:15--19, 1956.

\bibitem{pygraphviz}
\url{https://pygraphviz.github.io/}.

\bibitem{DBLP:conf/cp/RamaswamyS20}
Vaidyanathan~Peruvemba Ramaswamy and Stefan Szeider.
\newblock Maxsat-based postprocessing for treedepth.
\newblock In Helmut Simonis, editor, {\em Principles and Practice of Constraint
  Programming - 26th International Conference, {CP} 2020, Louvain-la-Neuve,
  Belgium, September 7-11, 2020, Proceedings}, volume 12333 of {\em Lecture
  Notes in Computer Science}, pages 478--495. Springer, 2020.
\newblock \href {https://doi.org/10.1007/978-3-030-58475-7\_28}
  {\path{doi:10.1007/978-3-030-58475-7\_28}}.

\bibitem{DBLP:conf/aaai/RamaswamyS21}
Vaidyanathan~Peruvemba Ramaswamy and Stefan Szeider.
\newblock Turbocharging treewidth-bounded bayesian network structure learning.
\newblock In {\em Thirty-Fifth {AAAI} Conference on Artificial Intelligence,
  {AAAI} 2021, Thirty-Third Conference on Innovative Applications of Artificial
  Intelligence, {IAAI} 2021, The Eleventh Symposium on Educational Advances in
  Artificial Intelligence, {EAAI} 2021, Virtual Event, February 2-9, 2021},
  pages 3895--3903. {AAAI} Press, 2021.
\newblock URL: \url{https://ojs.aaai.org/index.php/AAAI/article/view/16508}.

\bibitem{DBLP:conf/aaai/SchidlerS21}
Andr{\'{e}} Schidler and Stefan Szeider.
\newblock Sat-based decision tree learning for large data sets.
\newblock In {\em Thirty-Fifth {AAAI} Conference on Artificial Intelligence,
  {AAAI} 2021, Thirty-Third Conference on Innovative Applications of Artificial
  Intelligence, {IAAI} 2021, The Eleventh Symposium on Educational Advances in
  Artificial Intelligence, {EAAI} 2021, Virtual Event, February 2-9, 2021},
  pages 3904--3912. {AAAI} Press, 2021.
\newblock URL: \url{https://ojs.aaai.org/index.php/AAAI/article/view/16509}.

\bibitem{Sergeev2020}
Igor Sergeev.
\newblock On monotone circuit complexity of threshold boolean functions.
\newblock {\em Diskretnaya Matematika}, 32:81--109, 2020.
\newblock \href {https://doi.org/10.4213/dm1547} {\path{doi:10.4213/dm1547}}.

\bibitem{EPFLLibraries}
Mathias Soeken, Heinz Riener, Winston Haaswijk, Eleonora Testa, Bruno Schmitt,
  Giulia Meuli, Fereshte Mozafari, and Giovanni De~Micheli.
\newblock The {EPFL} logic synthesis libraries, November 2019.
\newblock arXiv:1805.05121v2.

\bibitem{DBLP:journals/mst/Stockmeyer77}
Larry~J. Stockmeyer.
\newblock On the combinational complexity of certain symmetric boolean
  functions.
\newblock {\em Mathematical Systems Theory}, 10:323--336, 1977.
\newblock \href {https://doi.org/10.1007/BF01683282}
  {\path{doi:10.1007/BF01683282}}.

\bibitem{DBLP:books/teu/Wegener87}
Ingo Wegener.
\newblock {\em The complexity of Boolean functions}.
\newblock Wiley-Teubner, 1987.
\newblock URL: \url{http://ls2-www.cs.uni-dortmund.de/monographs/bluebook/}.

\end{thebibliography}

\appendix
\section{Blocks for the Modulo~3 Function}
The following code justifies the existence of~circuits needed in the proof of Theorem~\ref{theorem:mod3upper}.
\begin{minted}{python}
from itertools import product

enc = {(0, 0): 0, (0, 1): 1, (1, 0): 2, (1, 1): 2}

# in_2
for x1, x2 in product(range(2), repeat=2):
    g1 = x1 ^ x2
    g2 = x1 & x2
    assert (x1 + x2) % 3 == enc[g2, g1]

# in_3
for x1, x2, x3 in product(range(2), repeat=3):
    g1 = x1 == x2
    g2 = 1 - (x1 | x2)
    g3 = g2 == x3
    g4 = g1 == g3
    g5 = g2 < g4
    assert (x1 + x2 + x3) % 3 == enc[g5, g3]

# in_4
for x1, x2, x3, x4 in product(range(2), repeat=4):
    g1 = x1 == x2
    g2 = g1 ^ x3
    g3 = g2 ^ x2
    g4 = g1 & g3
    g5 = g4 == x4
    g6 = g2 == g5
    g7 = g4 < g6
    assert (x1 + x2 + x3 + x4) % 3 == enc[g7, g5]

# mid_3
for x1, x2, x3, z0, z1 in product(range(2), repeat=5):
    g1 = x1 == z1
    g2 = g1 | z0
    g3 = g2 ^ x2
    g4 = g3 ^ z0
    g5 = g4 ^ x1
    g6 = g2 & g5
    g7 = g6 == x3
    g8 = g3 == g7
    g9 = g6 < g8
    assert (enc[z0, z1] + x1 + x2 + x3) % 3 == enc[g9, g7]

# out_1^1
for x1, z0, z1 in product(range(2), repeat=3):
    g1 = x1 ^ z1
    g2 = z0 < g1
    assert ((enc[z0, z1] + x1) % 3 == 1) == g2

# out_2^0
for x1, x2, z0, z1 in product(range(2), repeat=4):
    g1 = z0 == x2
    g2 = x1 ^ z1
    g3 = g1 == x1
    g4 = z0 < g2
    g5 = 1 - (g3 | g4)
    assert ((enc[z0, z1] + x1 + x2) % 3 == 0) == g5

# out_3^2
for x1, x2, x3, z0, z1 in product(range(2), repeat=5):
    g1 = x1 == z1
    g2 = g1 | z0
    g3 = g2 ^ x2
    g4 = g3 ^ z0
    g5 = g4 ^ x1
    g6 = g2 & g5
    g7 = g3 == x3
    g8 = 1 - (g6 | g7)
    assert ((enc[z0, z1] + x1 + x2 + x3) % 3 == 2) == g8

# mod3_3^0
for x1, x2, x3 in product(range(2), repeat=3):
    g1 = x2 == x3
    g2 = x1 ^ x2
    g3 = g1 > g2
    assert ((x1 + x2 + x3) % 3 == 0) == g3

# mod3_3^2
for x1, x2, x3 in product(range(2), repeat=3):
    g1 = x2 ^ x3
    g2 = x3 | g1
    g3 = x1 < g2
    g4 = g1 ^ g3
    assert ((x1 + x2 + x3) % 3 == 2) == g4

# mod3_4^2
for x1, x2, x3, x4 in product(range(2), repeat=4):
    g1 = x1 ^ x2
    g2 = x3 ^ x4
    g3 = x1 ^ x3
    g4 = g2 | g3
    g5 = g1 < g4
    g6 = g2 ^ g5
    assert ((x1 + x2 + x3 + x4) % 3 == 2) == g6
\end{minted}

\end{document}